\documentclass[11pt]{article}

\usepackage[final]{acl}

\usepackage{times}
\usepackage{latexsym}

\usepackage[T1]{fontenc}

\usepackage[utf8]{inputenc}

\usepackage{microtype}

\usepackage{inconsolata}

\usepackage{enumitem}
\usepackage{graphicx} 
\usepackage{amsmath} 
\usepackage{mathtools} 
\usepackage{booktabs}
\usepackage{CJKutf8}
\usepackage{makecell}
\usepackage{float}

\newcommand{\tensorconcat}{\mathbin{\oplus}}

\newcommand{\paratitle}[1]{\noindent \textbf{#1}}

\title{Harvesting Events from Multiple Sources: Towards a Cross-Document Event Extraction Paradigm}

\author{
  Qiang Gao, Zixiang Meng, Bobo Li, 
  Jun Zhou, Fei Li, Chong Teng, Donghong Ji\\ 
 Key Laboratory of Aerospace Information Security \\and Trusted Computing, Ministry of Education, \\School of Cyber Science and Engineering, Wuhan University \\
 \{qianggao, zixiangmeng, boboli, j$.$zhou, lifei\_csnlp, tengchong, dhji\}@whu.edu.cn
}

\begin{document}
\maketitle
\begin{abstract}
Document-level event extraction aims to extract structured event information from unstructured text. 
However, a single document often contains limited event information and the roles of different event arguments may be biased due to the influence of the information source.
This paper addresses the limitations of traditional document-level event extraction
by proposing the task of cross-document event extraction (CDEE) to integrate event information from multiple documents and provide a comprehensive perspective on events. 
We construct a novel cross-document event extraction dataset, namely CLES, 
which contains 20,059 documents and 37,688 mention-level events, where over 70\% of them are cross-document.
To build a benchmark, we propose a CDEE pipeline that includes 5 steps, namely event extraction, coreference resolution, entity normalization, role normalization and entity-role resolution. 
Our CDEE pipeline achieves about 72\% F1 in end-to-end cross-document event extraction, suggesting the challenge of this task.
Our work builds a new line of information extraction research and will attract new research attention.
Our code and dataset will be available at \href{https://github.com/cooper12121/CLES}{https://github.com/cooper12121/CLES}.

\end{abstract}

\section{Introduction}
In the realm of Natural Language Processing, document-level event extraction (\textbf{DEE}) has been a focal area of research, striving to distill structured information from unstructured text. 
This process typically involves identifying and categorizing events, along with their associated entities and relations, within a single document \citep{yang-etal-2018-dcfee}.  
This approach has demonstrated its effectiveness in numerous applications, such as information retrieval \citep{10.1145/3331184.3331415}, content summarization \citep{zhang2021event} and knowledge graph construction \citep{9792280}.

\begin{figure}[t]

\begin{center}
\includegraphics[width=\columnwidth]{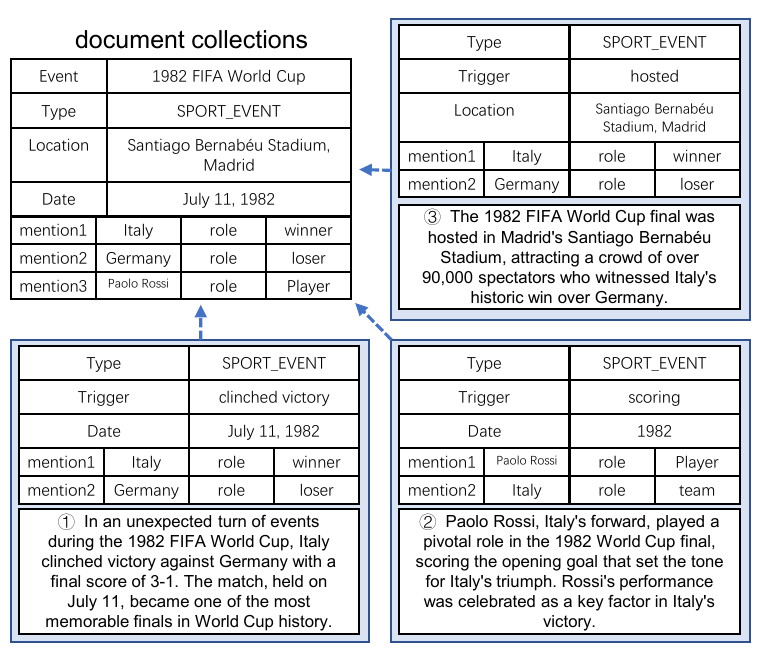} 
\caption{An example of cross-document event extraction, where a comprehensive event is obtained from three event mentions in three documents.
}
\label{fig.example}

\end{center}
\end{figure}

Although significant advancements have been made \cite{xu-etal-2021-document,yang-etal-2021-document,wang-etal-2023-document}, DEE often encounters limitations in terms of the scope and depth of information that it can provide.
Different documents may present varying perspectives or emphasize different aspects of the same event, leading to a fragmented and sometimes biased understanding when viewed in isolation.
Specifically, event information may be distributed across multiple documents. 
As shown in Figure \ref{fig.example}, three documents contain different event mentions referring to the same event, where the bottom-left document includes the ``Date'' argument while the top-right document includes the ``Location'' argument.


Recognizing these limitations, we propose the task of cross-document event extraction (\textbf{CDEE}),
which categorizes events into mention-level and concept-level. 
A mention-level event refers to the event defined within single document, while a concept-level event refers to a complete event obtained by integrating information from multiple documents.
Compared with DEE, 
The key issue that CDEE aims to address is the problem of completeness, which means obtaining a complete representation of an event by aggregating event information from multiple documents.
After integrating the extraction results of events from multiple documents, merging duplicate information and resolving conflicting information, the whole event can be built.

To foster research in this unexplored field, we have constructed a new cross-document event extraction dataset, called \textbf{CLES} (CrossLinkEventScope). 
Leveraging Wikipedia as the information source, we utilized the hyperlinks inside
to identify the documents relevant to events
and aggregate them into collections.
These collections not only encompass multiple perspectives of a single event but also include detailed background information and various viewpoints related to the event. 
Afterwards, we employed a DEE tool \citep{zhangHarvestText} to mine event mentions within documents and 
manually merge event mentions into a complete event.
The results in both processes were manually checked to guarantee the annotation quality.
Ultimately, the CLES dataset comprises 9 event types, over 37,688 mention-level events and 3,633 concept-level events, where over 70\% are cross-document.

Besides the dataset, we also contribute a CDEE pipeline comprising 5 steps:
(1) \textbf{DEE}: Event mentions as well as related arguments are extracted from individual document.
(2) \textbf{Event Coreference Resolution}: Event mentions within an event collection are grouped by coreference relations.
(3) \textbf{Entity Normalization}: A third-party entity linking library \citep{zhangHarvestText} is utilized to align entities and then their attributes are standardized.
(4) \textbf{Role Normalization}: The same type of roles across different documents are normalized by a role mapping table.
(5) \textbf{Entity-Role Resolution}: Extracted results from different documents are aligned and integrated to eliminate repetitive and conflicted content.

Our experimental results show that the CDEE pipeline is able to achieve about 72\% F1 for end-to-end cross-document event extraction, revealing its effectiveness but also the challenge of this task.
In the end, we highlight the contributions of this paper as:
\begin{enumerate}
    
    \item We introduce a novel CDEE task, aiming to extend the research scope of event extraction and provide a more comprehensive perspective.
    \item We construct a new large-scale CDEE dataset, which provides abundant data and lays the foundation for future research.
    \item We build a benchmark pipeline for CDEE, which can be used as a basic baseline for follow-up studies.
    
\end{enumerate}

\section{Related Work}

\subsection{Sentence-Level Event Extraction}
Sentence-level event extraction has been extensively researched \citep{liu-etal-2018-jointly-event,wadden-etal-2019-entity-event, Hamborg2019Giveme5W1HAU,wang-etal-2023-continual,xu-etal-2023-learning-friend}.
\citet{du-cardie-2020-event-qa} proposed a QA approach for event extraction to avoid the dependency of event extraction results on the previous entity recognition step. 
\citet{lu-etal-2021-text2event} adopted a seq2seq model for event extraction, transforming it into a test2event task. Compared to traditional methods, this approach avoids dividing event extraction into multiple subtasks and can yield results in a single step.
\citet{wang-etal-2022-deepstruct} introduced a novel structured pre-training framework that does not require fine-tuning on specific tasks. It transforms structured prediction into a sequence-based triple prediction task and achieved good results across multiple tasks.

\subsection{Document-Level Event Extraction}
In the past several years, there has been many methods and models for document-level event extraction, 
which can be categorized into several types: (1) Pipeline approach first identifies event triggers and event types, and then recognizes event arguments.
(2) Sequence labeling approach treats the task as a multi-class classification problem and directly performs sequence labeling on text sequences to identify event triggers and arguments.
(3) Graph-based methods and Generative-based methods.

\citet{yang-etal-2018-dcfee} transformed the DEE task into a Sentence-level Event Extraction (SEE) task, treating sentences containing event triggers and arguments as key-events. However, most of the dataset in this approach only detects information about a single event, without considering argument combinations.
\citet{zheng-etal-2019-doc2edag} addressed argument combinations by constructing a Directed Acyclic Graph (DAG) without relying on trigger words, named Doc2DAG. However, generating the DAG graph in this manner is heavily influenced by false positives and false negatives, and the computational overhead for building the graph is significant.

\citet{yang-etal-2021-document-DEPPN} employed a non-autoregressive decoder (NAD) and the Hungarian Algorithm for inference and gold matching. It achieves performance similar to Doc2EDAG with improved speed.
\citet{zhu2022efficient-PTPCG} explored a complete graph for event extraction, where all entity pairs of the same event are fully connected. However, since this method suffers from missing argument roles, pruning is introduced to alleviate this issue.

Considering the rich relational information among event parameters in documents, which can establish long-distance relationship knowledge for events, \citet{liang-etal-2022-raat} proposed a relation-enhanced document-level event extraction model.
Although this model has achieved significant improvements, relation prediction requires the introduction of an additional transformer framework, making the model more complex and increasing computational overhead.
\citet{wan-etal-2023-joint} introduced a Token-Token Bidirectional Event Completed Graph (TT-BECG) to addresse the inefficiency and error propagation problems associated with traditional pipeline methods.

\subsection{Cross-Document Information Extraction}
Although not many, there have been some studies on cross-document information extraction, 
such as event coreference resolution \citep{corefqa,focus-on,eirew-etal-2022-cross-search} and relation extraction \citep{yao-etal-2021-codred, lu-etal-2023-multi}.  

In terms of coreference resolution,
\citet{yu-etal-2022-pairwise} proposed a cross-document coreference resolution model that enhances event mention representation by extracting event arguments.
\citet{HGCN} addressed coreference resolution using a graph-based approach, while \citet{chen-etal-2023-cross-discourse} introduced discourse information to model documents, resulting in a significant performance improvement.
\citet{gao-etal-2024-enhancing-cross} proposed a cross-document coreference resolution model based on discourse information, modeling the structural and semantic information of documents through RST and lexical chains.

With respect to other directions,
\citet{caciularu-etal-2021-cdlm-cross} proposed a novel cross-document pre-training language model to learn rich contextual information across documents. \citet{wang-etal-2022-entity} proposed a cross-document relation extraction model based on bridge entities, which utilizes entity relation attention mechanisms across paths to facilitate interactions between entities.
To our knowledge, there are still no studies on cross-document event extraction.


\section{CLES: A Cross-Document Event Extraction Dataset}

\subsection{Objective Definition}
Our goal is to construct a large-scale, domain-agnostic cross-document event extraction dataset, which covers a wide range of event types to reflect the rich content and diversity of Wikipedia. Additionally, we do not set restrictions on the time span of events, allowing for the inclusion of historical and contemporary events to enhance the temporal dimension and depth. 
Moreover, we select Chinese as the main language in building this dataset.
To ensure the diversity and comprehensiveness of the dataset, we have defined a total of nine event categories, including ATTACK EVENT, SPORT EVENT, EVENT UNK, ELECTION EVENT, GENERAL EVENT, DISASTER EVENT, ACCIDENT EVENT, AWARD VENT, and OTHERS. 


In constructing the dataset for cross-document event extraction, our main idea is to leverage Wikipedia as the information source and utilize hyperlinks added by authors when creating articles to identify and aggregate documents related to events. Each Wikipedia article typically pertains to a specific topic or event, and authors often add hyperlinks to key phrases that point to other related articles or detailed pages about events. These hyperlinks naturally form a network, clustering different documents together based on events.

\begin{figure}[t]

\begin{center}
\includegraphics[width=\columnwidth]{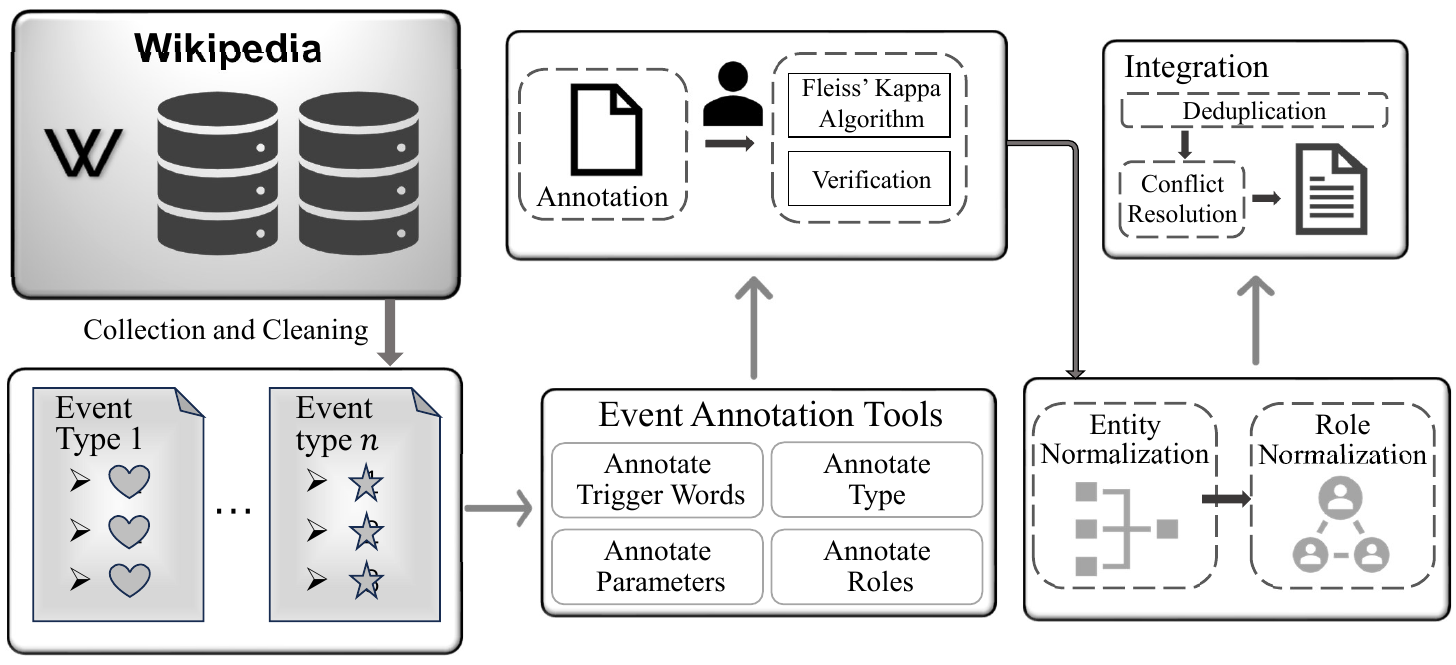} 
\caption{The process of dataset construction.
}
\label{fig.process}

\end{center}
\end{figure}

Using this hyperlink network, we can cluster all documents pointing to the same event or topic, forming the collections of articles centered around specific events. These collections not only encompass multiple perspectives on a single event but also include detailed background information and various viewpoints related to the event. By analyzing and integrating these documents, we can capture comprehensive information about events from multiple sources and perspectives, providing a rich and multidimensional data foundation for cross-document event extraction.
The process of dataset construction is illustrated in Figure \ref{fig.process} and the details in the process are explained in the following sections.

\subsection{Data Collection and Cleaning}
We borrowed the approach from \citet{eirew-etal-2021-wec} and optimized its data collection system to gather data from Wikipedia dump files.
After obtaining the raw data, we first conducted data cleaning to ensure the quality and relevance to our goal, which involved several substeps.
(1) Removing Non-Event Documents: We reviewed all crawled documents and excluded those that did not clearly describe specific events. For example, some documents might only briefly mention an event without providing detailed information or background about it. 

(2) Filtering Unrelated Documents: We filtered out documents that were unrelated to the current event document collection. Only the documents directly related to the events were retained in the dataset to ensure consistency and accuracy of the data.

\subsection{Annotation and Validation}
After completing data cleaning, we carried out a data selection process, that is, the maximum number of documents in each document collection is 10. 
For the document collections with more than 10 documents, we manually selected 10 documents with the richest event information.

Due to the large scale of documents in our data, the cost of manual annotation for all documents is prohibitively high. 
Therefore, we used an event extraction tool for annotating each document and then conducted manual verification. We adopted the method proposed by \citet{peng2023devil} to label event trigger words, event arguments and their roles. 
To ensure the quality of the dataset, two annotators independently verified the results annotated by the tool and corrected labeling errors. We calculated the consistent rate between the tool and human annotators using the Fleiss’ Kappa algorithm \citep{Kappa}.
The kappa value is 0.72, indicating decent 
annotation quality of our dataset.

Based on the single-document event information annotated in the previous step, annotators de-duplicated the event-related argument information. Additionally, they eliminated irrelevant events based on the original document information. In the cases where an entity was assigned with multiple roles, the most accurate role was selected based on context. Ultimately, this process yielded the final event for each document collection.
More specifically, in this process, based on our constructed role table and existing entity linking tools, we first perform coarse-grained filtering through a program we wrote, followed by verification and refinement of the merged results by annotators.
\begin{table}[t]
\centering
\resizebox{0.5\textwidth}{!}{
\renewcommand{\arraystretch}{1.25}

\begin{tabular}{lcccc}
\hline
 & {Docs} & \makecell{Mention-level \\Events} & 
 \makecell{Concept-level\\ Events} & \makecell{Cross-document\\ Events (\%)}\\
\hline
{Train} & {17,163} & 32,311 & {3,855}   &{71.2\%} \\
{Dev}   & {1,387}  & 2,540  & {297}       &{71.7\%}\\
{Test}  & {1,509}  & 2,817  & {324}       &{76.5\%}\\
\hline
{All} & {20,059} &{37,668}  &{4,476}        &{71.6\%}\\
\hline
\end{tabular}
}
\caption{\label{table.1}
The statistics of documents and events in CLES.
Mention-level events refers to the events annotated within documents, and concept-level event represents the events merged from multiple documents in the collection.
}

\end{table}

\begin{table}[t]
\centering
{\fontsize{9}{10}\selectfont
\renewcommand{\arraystretch}{1.25}
\begin{tabular}{lccc}
\hline
 & {Train} & {Dev} & {Test} \\
\hline
{ATTACK EVENT      }    & {10,156} & {785} & {804}\\
{SPORT EVENT       }     & {2,370}  & {140} & {246}\\ 
{EVENT UNK}       & {1,580}  & {127} & {110}\\
{ELECTION EVENT}  & {1,323}  & {113} & {146} \\
{GENERAL EVENT}   & {758}    & {138} & {127}\\
{DISASTER EVENT}  & {261}    & {31} & {39}\\
{ACCIDENT EVENT}  & {352}    & {39} & {20}\\
{AWARD EVENT}     & {105}    & {12} & {15}\\
{OTHERS}          & {158}    & {2} & {2}\\
\hline
\end{tabular}
\caption{\label{table.Event_types}
The number of documents for each event type.
}
}
\end{table}

\begin{table}[t]
\centering
\begin{tabular}{lccc}
\hline
 & {Train} & {Dev} & {Test} \\
\hline

{documents=1} & {1,110} & {84} & {76}\\
{documents=2} & {528}   & {53} & {76}\\ 
{documents=3} & {444}   & {5}  & {10}\\
{documents=4} & {296}   & {17} &{18} \\
{documents=5} & {210}   & {24} & {25}\\
{documents=6} & {133}   & {22} & {11}\\
{documents=7} & {121}   & {10} & {17}\\
{documents=8} & {124}   & {10} & {16}\\
{documents=9} & {96}    & {8}  & {9}\\
{documents=10}& {793}   & {64} & {66}\\

\hline
\end{tabular}
\caption{\label{table.collections}
The numbers of document collections with respect to collection sizes.
}
\end{table}

\begin{table}[!ht]
\centering
\renewcommand{\arraystretch}{1.25}
\resizebox{0.5\textwidth}{!}{
\begin{tabular}{l|ccc}
\hline
 & {Train} & {Dev} & {Test} \\
\hline
{doc min length}                & {15}        & {12}      & {15}\\
{doc avg length}                & {210.1}     & {206.3}   & {197.5} \\
{doc max length}                & {4,553}     & {1,626}   & {1,416}\\
\hline
{trigger number}                 & {32,311}    & {2,540}   & {2,817}\\
{trigger avg length}            & {2.06}      & {2.07}    & {2.06}\\
{trigger avg number per doc}    & {1.88}      & {1.83}    & {1.87}\\
\hline
{role number}                    & {81,270}    & {6,231}   & {6,848}\\
{role avg number per event}     & {2.52}      & {2.45}    & {2.43}\\
{unique role number}     & {469}      & {136}    & {157}\\
\hline

\hline
\end{tabular}
}
\caption{\label{table.length}Statistics related to trigger words, argument roles, and their lengths. All lengths refer to the numbers of words. 
}
\end{table}

\subsection{Dataset Statistical Analysis}\label{dataset analysis}

The scale of the final dataset is shown in Table \ref{table.1}. 
In terms of scale, our dataset consists of over 20,000 documents and 37,000 events. This demonstrates that our dataset covers a vast amount of event information, spanning a wide range of time frames and diverse textual content. This necessitates event extraction models to possess strong generalization capabilities. Furthermore, the proportion of cross-document events in our dataset exceeds 70\%, indicating that the majority of events require synthesizing information from multiple documents, posing a challenge to the modeling capacity.

We also count the number of documents for each event type as shown in Table \ref{table.Event_types}.
It can be observed that our dataset has a power-law distribution across different event types, with ATTACK events being the most common. This also indicates that Wikipedia has the highest number of articles related to attack events. 

To delve into the distribution of document collection sizes, we compiled the statistics on the distribution of the document number in each document collection, as shown in Table \ref{table.collections}.
It can be seen that our dataset has reasonable distributions of different document collection sizes. It contains both cross-document events and a certain number of single-document events. This indicates that even in the context of cross-document extraction, there are still some events that can be fully extracted from a single document. 
Therefore, our dataset can be used to evaluate the methods not only for document-level event extraction but also for cross-document event extraction.

Moreover, the statistical information related to trigger words, argument roles, and their lengths can be found in Table \ref{table.length}.
The presence of long documents necessitates the model capability of handling long-distance dependency in text context and events. 
The average of 1.8 trigger words per document indicates that there may be multiple events within single document, posing a challenge for event extraction models. 
The total number of unique roles is 469, suggesting good uniformity in role definitions within our dataset.
Other details of the dataset can be found in Appendix \ref{sec:appendix dataset}.

\begin{figure*}[!h]
\begin{center}
\includegraphics[width=\textwidth,scale=1]{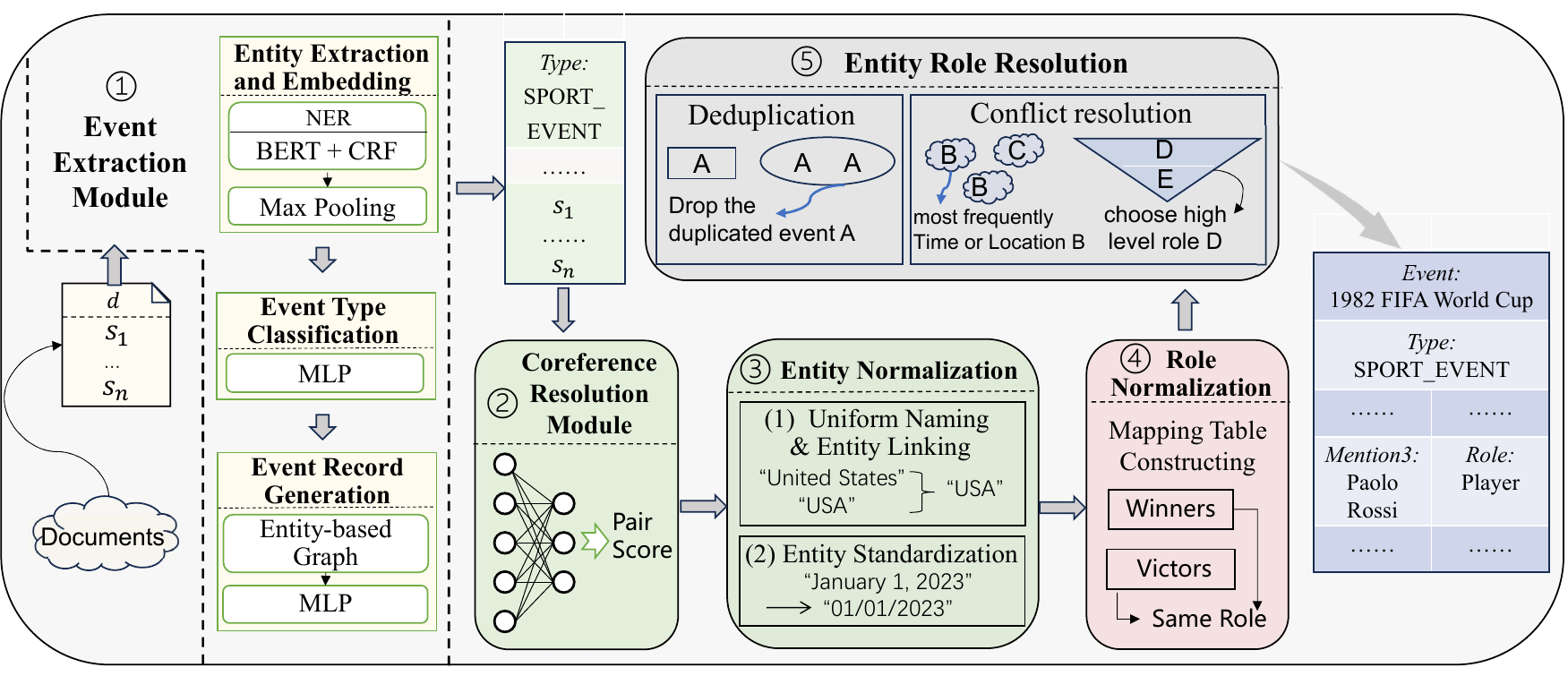} 
\caption{
Our CDEE pipeline framework consists of five main components. 
Event Extraction Module performs document-level event extraction. $d$ represents the document, and $s_i$ represents the $i$-th sentence in the document.
Event Coreference Resolution Module clusters the event mentions of the same event.
Entity Normalization Module links entities to a knowledge base.
Role Normalization Module unifies the descriptions of event argument roles.
Entity-Role Resolution Module performs deduplication and conflict resolution for cross-document events in the document collection.
}
\label{fig.2}
\end{center}
\end{figure*}

\section{Our Pipeline Framework for CDEE}
Based on our dataset, we propose a new cross-document event extraction framework, which mainly consists of the following components: event extraction, coreference resolution, entity normalization, role normalization and entity-role resolution. 

The framework architecture is illustrated in Figure \ref{fig.2}.
First, the documents in a collection are input into the event extraction module and output independent event extraction results for each document. 
Then, the coreference resolution module eliminates irrelevant events and 
clusters the event mentions of the same event.
Subsequently, event arguments are normalized by the entity and role normalization modules respectively.
Finally, the entity-role resolution module 
performs deduplication and conflict resolution for the cross-document event extraction results, and yields complete representations of the events in the document collection.

\subsection{Document-level Event Extraction}\label{event extraction module}

We follow the approach of \citet{zheng-etal-2019-doc2edag} to construct an entity-based directed acyclic graph (EDAG) from the event records to perform event extraction. 
The model is mainly divided into the following parts:

(1) Entity Extraction and Embedding: Named Entity Recognition (NER) is performed using BERT \citep{devlin2019bert} and CRF to obtain the embeddings of entities and sentences.

(2) Document-level Encoding: Transformer encoder and RST (Rhetorical Structure Theory) \citep{RST} tree are used to make entities aware of document-level context. 

(3) Event Type Classification: Event classification task is performed based on the document-level sentence representations obtained from the previous step.

(4) Event Role Extraction: A directed graph based on entities is constructed to determine the event roles between entities.

The specific details of the event extraction model are provided in Appendix \ref{appendix:deatil of event extraction}.

\subsection{Event Coreference Resolution}
To ensure that all events in the document collection refer to the same event, without including any other irrelevant noise, 
following \citet{gao-etal-2024-enhancing-cross} ,we introduce a coreference resolution module to group the event collections belonging to the same event. 
Here, building upon the event representations enhanced with RST obtained in Section \ref{event extraction module},
we train a multi-layer MLP to serve as the coreference resolution model, and perform binary classification to determine whether two event mentions are co-referential. 
For two event mentions \( e_i \) and \( e_j \) output by the event extraction module, the coreference probability is calculated by: 
\begin{equation}
 p = \text{MLP}(m_i, m_j, m_i \cdot m_j) 
\end{equation}
where \( m_i \) and \( m_j \) are the vector representations of events \( e_i \) and \( e_j \) respectively.
In our method, we compute coreference scores for all event pairs within the same document collection. We retain all event mentions with the coreference scores greater than 0.5 for subsequent steps.

\subsection{Entity Normalization}
We normalize the extracted entities by linking them to a knowledge base to ensure consistent representation of the same entity across different documents. We use the method proposed by \citet{zhangHarvestText} as our entity linking module.
This method integrates existing Chinese dictionaries and utilizes contextual information for word sense disambiguation. 
Entity normalization mainly involves the following steps:

(1) Entity Linking: Linking entities in the document to the corresponding entities in the knowledge base to address the issues of homonyms and synonyms. For example, ``United States'' and ``USA'' should be represented as the same entity.

(2) Entity Standardization: Standardizing various attributes of entities, such as unifying date formats (e.g., ``January 1, 2023'' and ``01/01/2023'' to a standard format) and standardizing location names.
This can be implemented using existing time standardization tools.

\subsection{Role Normalization}

This module ensures that the terms describing the same type of event argument roles are consistent across different documents. For example, if ``winner'' and ``victors'' refer to the same type of roles in different documents, they should be normalized. 
Based on the specific information of the roles that we gathered in Section \ref{dataset analysis}, we manually designed a mapping dictionary for roles to ensure that all roles appearing in our dataset can be mapped to unified representations.
Our dataset contains a total of 469 unique roles. 
Based on this, we collected existing roles to build a role-mapping table. 
During the normalization process, if a role is not in the role mapping table, we add it as a new role to the mapping table. Table \ref{table.roles} and  Table \ref{table.roles_low} provide some examples.

\subsection{Entity-Role Resolution}
The entity-role resolution module aligns and consolidates the results of multi-document event extraction. This module performs two main operations: deduplication and conflict resolution.

\paratitle{Deduplication}, which ensures that only one instance of duplicated event mentions extracted from multiple documents is retained.

\paratitle{Conflict resolution}
   (1) Conflict resolution in event time and location arguments: We select the time and location arguments that appear most frequently across the documents.
   (2) Conflict resolution in argument roles: When the same entity is assigned with different roles, we resolve such conflict by constructing a hierarchical role selection mechanism, where the high-level role is selected. For detailed information about the hierarchical role selection mechanism, refer to Appendix \ref{role hierarchy}.
   
\section{Experiments}
Given that our proposed model follows a pipeline architecture, we designed three experiments to test the performance of each module: document-level event extraction experiment, event coreference resolution experiment, and cross-document event extraction experiment. These three experiments precisely reflect the three most important aspects of our framework. Through the document-level event extraction experiment, we can validate the effectiveness of incorporating RST. Through the event coreference resolution experiment, we can evaluate the accuracy of our coreference resolution module in removing irrelevant events. Through the cross-document event extraction experiment, we are able to show the effectiveness of our pipeline framework.

\subsection{Document-Level Event Extraction}\label{sec:extraction experiment}

\paratitle{Baselines} We choose Doc2EDAG \citep{zheng-etal-2019-doc2edag} 
as the baseline. Since Doc2EDAG can only be used for document-level event extraction, we compared our document-level event extraction module with this approach to demonstrate the impact of introducing discourse-level information (e.g., RST). We also compared our method with the approach RAAT \citep{liang-etal-2022-raat}, which incorporates additional entity relations to enhance the model performance.

\paratitle{Metrics} We use recall, precision, and $\text{F}_\text{1}$ score as evaluation metrics. We separately calculate the metrics for the tasks of event type classification and event role extraction.

\paratitle{Results}
The experimental results are shown in Table \ref{table.extraction}.
Compared to Doc2EDAG, our module achieved an F1 score improvement of 0.4 for event type classification and 4.6 for event role extraction, respectively. This is because we introduced RST to better model document information, and GAT can learn rich structural information contained in the RST tree. 
For the event type classification task, our dataset is domain-independent and only includes 9 major event types, making it less challenging. However, the event role extraction task requires rich document information. Our RST tree can provide rhetorical relationships between different clauses in the document, helping the model filter out noise.
Compared to RAAT, it does not have any particular features, hence the results of event type classification are similar. However, for event role extraction, RAAT introduces additional entity relationship information, leading to a noticeable improvement. RAAT results are close to ours, indicating that both entity relations and discourse information can enhance document understanding of the model.

\begin{table*}[h]
    \centering
    {\fontsize{4}{4}\selectfont
    \renewcommand{\arraystretch}{1.25}
    \resizebox{\textwidth}{!}{
    \begin{tabular}{@{}lc ccccccc@{}}
    \hline
    & \phantom{abc}&\multicolumn{3}{c}{Event Type Classification} & \phantom{abc}& \multicolumn{3}{c}{Event Role Extraction} \\
    \cmidrule{3-5} \cmidrule{7-9}
    && R & P & $\text{F}_\text{1}$ && R & P & $\text{F}_\text{1}$  \\ 
    \hline
        Doc2EDAG \citep{zheng-etal-2019-doc2edag} && 
        87.9 & 84.2 & 86.0 &&
         72.9 & 76.2 & 74.5
         \\
        RAAT \citep{liang-etal-2022-raat} && 87.6 & 84.7 & 86.1 &&  75.6 & 81.5  & 78.4  \\
        Our event extraction module && 87.8 & 85.1 & 86.4 && 76.9 & 81.4 & 79.1  \\
    \hline
    \end{tabular}
    }
    \caption{The results of document-level event extraction.
    }
    \label{table.extraction}
    }
\end{table*}

\begin{table*}[!t]
    \centering
    \resizebox{\textwidth}{!}{
    \begin{tabular}{@{}lccc ccc ccc c cccc@{}}
    \hline
    & \phantom{abc}&\multicolumn{3}{c}{MUC} & \phantom{abc}& \multicolumn{3}{c}{$\text{B}^\text{3}$} & \phantom{abc}& \multicolumn{3}{c}{CEAF} & \phantom{abc}& CoNLL\\
    \cmidrule{3-5} \cmidrule{7-9} \cmidrule{11-13} \cmidrule{15-15}
    && R & P & $\text{F}_\text{1}$ && R & P & $\text{F}_\text{1}$ && R &P & $\text{F}_\text{1}$ && $\text{F}_\text{1}$  \\ 
    \midrule
        \citet{yu-etal-2022-pairwise} &&
        79.2 & 83.4 & 81.2 && 
        81.3 & 78.5 & 80.2 && 
        77.6 & 81.7 & 79.6 && 
        80.3 \\
        Our coreference-\\ resolution module && 
        82.9 & 85.6 & 84.2 && 
        85.4 & 80.1 & 82.7 && 
        76.3 & 82.8 & 79.4 && 
        82.1 \\
    \bottomrule
    \end{tabular}
    }
    \caption{
    The results of event coreference resolution. 
    We conducted experiments on the event mentions generated from Section \ref{sec:extraction experiment}.
    }
    \label{table.coreference resolution}
\end{table*}

\begin{table*}[!t]
    \centering
    \small\
    {\fontsize{4}{4}\selectfont
    \renewcommand{\arraystretch}{1.25}
    \resizebox{0.6\textwidth}{!}{
    \begin{tabular}{lccc}
    \hline
    & R & P & $\text{F}_\text{1}$ \\
    \hline
        Baseline &
          71.3 & 68.2  & 69.7  \\

        Our pipeline framework &
           74.8 & 70.5 &72.6  \\      
        Llama2-Chinese-7b-Chat & 80.2& 78.1 & 79.1 \\
    \hline
    \end{tabular}
    }
    \caption{The results of cross-document event extraction. 
    }
    \label{table.cross event extraction}
    }
\end{table*}

\subsection{Event Coreference Resolution}
To remove event information irrelevant to the theme events of the document collection, we conducted cross-document event coreference resolution experiments. 

\paratitle{Baseline} We chose \citet{yu-etal-2022-pairwise} as the baseline, which determines coreference by enhancing event mention representations with event argument information. 

\paratitle{Metrics} We utilized the agglomerative clustering algorithm for clustering and reported R, P, and $\text{F}_\text{1}$ scores on the MUC, $\text{B}^\text{3}$, CEAF, and CoNLL metrics.

\paratitle{Results} The experimental results are shown in Table \ref{table.coreference resolution}.
Our model achieved certain performance improvements across all metrics. This is because there is often irrelevant event argument information  at the document level. Extracting arguments such as locations and times may result in incorrect results, and the event arguments in documents often exhibit long-distance dependency phenomena. Thus, it is necessary to explicitly introduce the structural information of the document, distinguishing the roles of different clauses. The RST tree used in our document encoding module is able to alleviate this issue.

\subsection{Cross-document Event Extraction}

\paratitle{Baseline} Since there are no existing works in the field of cross-document event extraction, we designed a rule-based baseline for comparison. 
This baseline operates on our coreference resolution results, and performs entity/role normalization and integration. 
For entity/role normalization, we used dictionary matching to standardize entities and roles. 
For role integration, we employed the principle of maximum count, selecting the role with the highest frequency for each entity as the final entity role. 
The details of this baseline can be found in our code.

\paratitle{Metrics} We use recall, precision and $\text{F}_\text{1}$ as evaluation metrics for Event Type Classification task and Event Role Extraction task. For Corefercence resolution, we use MUC, B$^3$, CEAF and CoNLL as evaluation metrics.

\paratitle{Results}
The experimental results are shown in Table \ref{table.cross event extraction}.
It can be observed that our method outperforms the rule-based baseline. This is due to the fact that the rule-based baseline does not consider the context of entity occurrences during entity/role normalization, leading to more errors. Additionally, constructing role hierarchies helps resolve conflicts and is superior to the maximum count method.

\subsection{Experiments Using LLM}
To further show the challenge of our dataset and the complexity of our task, we conducted additional experiments using LLMs.
The choice to employ LLM in our experiments stems from their advanced capabilities in handling various NLP tasks, which are essential for tackling the intricate challenges presented by our dataset:
\begin{enumerate}
    \item Cross-document context demands a model with well understanding capability of long text and different topics.
    
    \item Complex task procedure asks a model for the ability of assembling various information extraction skills such as trigger extraction, entity normalization and conflict resolution.
   
\end{enumerate}

\paratitle{Settings}
We used Llama2-Chinese-7b-Chat\footnote{https://github.com/LlamaFamily/Llama-Chinese} to finetune on our dataset with 4 A100-80G GPUs. 
The learning rate is set to 2e-6, and the batch size is set to 8.
The input prompt template is show in 
Textbox \textbf{Input Prompt}.

% ...

'''
To accomplish the cross-document event extraction task, you will be provided with multiple documents. Your objective is to extract event information from these documents, integrate the extracted results for multiple events, and perform entity and role normalization during the integration process. This involves linking entities and roles to a unified representation, while filtering out irrelevant event extraction results. Subsequently, you will merge multiple results into a comprehensive structured representation of events. The output format should be as follows:
\{
    "type": event type, 
    "trigger": event trigger,
    "arguments": [{"role": role1, "entity": entity1},{"role": role2, "entity": entity2}],
    …… 
\}\\
document\_input: \\
document1: \{....\},\\
document2: \{....\},\\
...
'''

\paratitle{Results}
The experimental results are shown in Table \ref{table.cross event extraction}. Note that for LLM we have a more flexible approach for evaluation metrics calculation, that is, if the predicted outputs are contained within the gold standards, we consider the result to be correct.

The experimental results show that the use of LLM leads to significant performance improvements. This is because the LLM has been pre-trained on a large amount of general data, possessing substantial knowledge capabilities. Furthermore, our dataset is domain-agnostic and covers a wide range, which contributes to the good performance of the model.
Additionally, we have found that using fully parameterized finetuning tends to overfit our task. 
Although the training loss decreases, the error rate is relatively high in the test set, especially when we increase the number of documents per event. 

Furthermore, the model's outputs are greatly influenced by the prompts. 
We observed that when limiting the model to output results in JSON format, it does not always comply as expected. 
Also, our prompts are constructed entirely in a zero-shot manner, where all sample labels are in JSON format, yet the model does not always adhere to our specifications. 
Moving forward, we plan to explore training the LLM using a few-shot approach to see if we can further improve performance.

\section{Conclusion}
In this paper, we introduced a novel task of cross-document event extraction.
A large-scale dataset, CLES, is proposed based on Wikipedia and a benchmark pipeline is built for the comparison of follow-up work.
Experiments show the feasibility and challenges of our task and dataset.
Our work paves the way for a more complex and comprehensive understanding of events, highlighting the importance of multi-document analysis in capturing real-world events.
Our work extends the scope of information extraction and will lead a new line of NLP research.

\bibliography{custom}

\appendix

\section{CLES Dataset}
\label{sec:appendix dataset}
To provide a more detailed overview of our dataset, we present some statistical information about the dataset here.

To further analyze the distribution of trigger words and roles in our dataset, 
we have compiled the statistics for the trigger words with the highest and lowest frequencies of occurrence, as shown in Tables \ref{table.triggers} and \ref{table.triggers_low_freq}, respectively. We also have compiled
the statistics for the roles with the highest and lowest frequencies of occurrence, as shown in Tables \ref{table.roles} and \ref{table.roles_low}. 

It can be observed that ``obtain’‘ and ``defeat" appear most frequently as trigger words, which is related to the fact that ATTACK type events are most common in our dataset. Additionally, ``Date'' and `Location'' appear most frequently as event arguments, indicating that Date and location information are often essential arguments for events.

\begin{table*}[!ht]
    \centering
    {\fontsize{4}{5}\selectfont
    \renewcommand{\arraystretch}{1.5}
    \resizebox{\textwidth}{!}{
    \begin{tabular}{@{}cc ccc ccc@{}}
    \hline
    \multicolumn{2}{c}{Train} & \phantom{abc}& \multicolumn{2}{c}{Dev}  &\phantom{abc}& \multicolumn{2}{c}{Test} \\
    \cmidrule{1-2} \cmidrule{4-5} \cmidrule{7-8}
    {trigger}&{numbers} && {trigger}&{numbers} && {trigger}&{numbers} \\
    \hline
    {defeat}&{656}  && {obtain}&{57} && {obtain}&{77}\\
    {obtain}&{481}  && {defeat}&{45} && {defeat}&{57}\\
    {demand}&{417}  && {demand}&{43} && {die}&{31}\\
    {command}&{373} && {organize}&{33} && {demand}&{28}\\
    {invade}&{338}  && {conflict}&{33} && {champion}&{26}\\
    {organize}&{335}&& {die}&{30} && {command}&{26}\\
    {conflict}&{317} && {invade}&{29} && {organize}&{23}\\
    {occupy}&{304}  && {command}&{27} && {conflict}&{22}\\
    {champion}&{288} && {support}&{26} && {occupt}&{22}\\
    {support}&{286}  && {surrender}&{20} && {abandon}&{19}\\
    \hline
    \end{tabular}
    }
    \caption{
    High frequency event trigger word statistics, where ``number'' indicates the frequency of occurrence for each trigger word.
    }
    \label{table.triggers}
    }
\end{table*}
\begin{table*}[!ht]
    \centering
    {\fontsize{4}{5}\selectfont
    \renewcommand{\arraystretch}{1.5}
    \resizebox{\textwidth}{!}{
    \begin{tabular}{@{}cc ccc ccc@{}}
    \hline
    \multicolumn{2}{c}{Train} & \phantom{abc}& \multicolumn{2}{c}{Dev}  &\phantom{abc}& \multicolumn{2}{c}{Test} \\
    \cmidrule{1-2} \cmidrule{4-5} \cmidrule{7-8}
    {trigger}&{numbers} && {trigger}&{numbers} && {trigger}&{numbers} \\
    \hline
    {follow}&{1}  && {flee}&{1} && {abdicate}&{1}\\
    {disguise}&{1}  && {withdraw}&{1} && {visit}&{1}\\
    {succeed}&{1}  && {cease}&{1} && {warn}&{1}\\
    {disappear}&{1} && {resale}&{1} && {impeach}&{1}\\
    {steal}&{1}  && {relocate}&{1} && {negotiate}&{1}\\
    {launch}&{1}&& {bind}&{1} && {ambush}&{1}\\
    {burn}&{1} && {fail}&{1} && {dissolve}&{1}\\
    {crush}&{1}  && {coup}&{1} && {rescue}&{1}\\
    {refund}&{1} && {assassinate}&{1} && {divide}&{1}\\
    {delete}&{1}  && {debate}&{1} && {repair}&{1}\\
    \hline
    \end{tabular}
    }
    \caption{
    Low frequency event trigger word statistics, where ``number'' indicates the frequency of occurrence for each trigger word.
    }
    \label{table.triggers_low_freq}
    }
\end{table*}

\begin{table*}[!ht]
    \centering
    {\fontsize{6}{7}\selectfont
    \renewcommand{\arraystretch}{1.45}
    \resizebox{\textwidth}{!}{
    \begin{tabular}{@{}cc ccc ccc@{}}
    \hline
    \multicolumn{2}{c}{Train} & \phantom{abc}& \multicolumn{2}{c}{Dev}  &\phantom{abc}& \multicolumn{2}{c}{Test} \\
    \cmidrule{1-2} \cmidrule{4-5} \cmidrule{7-8}
    {role}&{numbers} && {role}&{numbers} && {role}&{numbers} \\
    \hline
    {date}&{23,403}                   && {date}&{1,812}          && {date}&{1,942}\\
    {location}&{6,834}                && {location}&{555}         && {location}&{560}\\
    {attacker}&{6,631}                && {attacker}&{484}         && {attacker}&{475}\\
    {winner}&{5,714}                  && {target}&{417}           && {loser}&{441}\\
    {loser}&{5,699}                   && {winner}&{374}           && {winner}&{440}\\
    {target}&{5,553}                  && {loser}&{372}            && {target}&{414}\\
    {victim}&{3,323}                  && {victim}&{340}           && {victim}&{246}\\
    {competition}&{1,934}             && {competition}&{121}       && {champion}&{160}\\
    {organization}&{1,612}            && {award}&{117}            && {competition}&{146}\\
    {champion}&{1,583}                && {recipient}&{116}        && {championship}&{145}\\
    \hline
    \end{tabular}
    }
    \caption{
    High frequency event role statistics, where ``number'' indicates the frequency of occurrence for each role.
    }
    \label{table.roles}
    }
\end{table*}
\begin{table*}[!ht]
    \centering
    {\fontsize{6}{7}\selectfont
    \renewcommand{\arraystretch}{1.45}
    \resizebox{\textwidth}{!}{
    \begin{tabular}{@{}cc ccc ccc@{}}
    \hline
    \multicolumn{2}{c}{Train} & \phantom{abc}& \multicolumn{2}{c}{Dev}  &\phantom{abc}& \multicolumn{2}{c}{Test} \\
    \cmidrule{1-2} \cmidrule{4-5} \cmidrule{7-8}
    {role}&{numbers} && {role}&{numbers} && {role}&{numbers} \\
    \hline
    {destroyer}&{1}       && {decrypter}&{1}          && {straying party}&{1}\\
    {translator}&{1}     && {fined entity}&{1}      && {strayed individual}&{1}\\
    {exporter}&{1}      && {enforcement authority}&{1}   && {provider}&{1}\\
    {comforter}&{1}    && {evacuating party}&{1}   && {independent party}&{1}\\
    {pollutant}&{1}       && {candidate}&{1}   && { terminating party}&{1}\\
    {owner}&{1}            && {member}&{1}        && {resigning party}&{1}\\
    {declarant}&{1}      && { leader}&{1}           && {transaction}&{1}\\
    {leader}&{1}        && {warring party}&{1}    && {practitioner}&{2}\\
    {provider}&{1}       && {recipient}&{1}           && {commander}&{2}\\
    {issuer}&{1}          && {occupier}&{2}        && {acquiring party}&{2}\\
    \hline
    \end{tabular}
    }
    \caption{
    Low frequency event role statistics, where ``number'' indicates the frequency of occurrence for each role.
    }
    \label{table.roles_low}
    }
\end{table*}

\section{Cross Document Event Extraction Architecture}\label{cross-model}

\subsection{The Details of Document-Level Event Extraction}\label{appendix:deatil of event extraction}
\subsubsection{Entity Extraction and embedding}
Firstly, we need to perform Named Entity Recognition (NER) on the input document. We employ a sota model proposed by \citet{wang2022improving}, which utilizes BERT \citep{devlin2019bert} with a Conditional Random Field (CRF) layer for NER. Given a document \( d = \{s_1, s_2, ..., s_n\} \), where \( s_i \) represents a sentence, NER processing yields an entity set \( E = \{e_1, e_2, e_3, ..., e_j\} \), where \( e_i \) represents an entity.
Since an entity mention may consist of multiple tokens, we employ the maximum pooling result of these tokens as the embedding for the entity mention, i.e., \( e_i = \text{max-pooling}([h_{i,j}, ..., h_{i,k}]) \), where \( h_{i,j} \) represents the representation of the \( j \)-th token of mention \( i \). For each sentence \( s_i \), we also adopt the maximum pooling method to obtain the embedding of each sentence, \( c_i = [h_{i,1}, ..., h_{i,n}] \), where \( h_{i,j} \) represents the token of the \( j \)-th token of sentence \( s_i \).

\subsubsection{Document-level Encoding}
In the previous section, we obtained embeddings for each entity and sentence, encoding only the contextual information within the sentence scope. However, without interaction among the sentences of the document, this local encoding may not be sufficient for direct event parameter extraction, as the event parameter information may be distributed across different sentences. Therefore, it's necessary to make entities and sentences aware of the document-level context. To encode document information more effectively, we introduce discourse information to model document information. We construct an RST tree to represent the rhetorical relations between different clauses in the document. The document is divided into Elementary Discourse Units (EDUs), and the constructed RST is used for subsequent processing. We use two modules to learn document-level context information:

(1) Transformer Encoder: The entity embeddings and sentence embeddings obtained from section 4.1 are added with positional encodings and then fed into the transformer encoder for interaction between different entities and sentences. \( E_t = [e_t^1, ..., e_t^{N_e}] = \text{transformer}(e_1, ..., e_{N_e}, c_1, ..., c_{N_s}) \) + position encoding, and \( C_t = [c_t^1, ..., c_t^{N_s} ]= \text{transformer}(e_1, ..., e_{N_e}, c_1, ..., c_{N_s}) \) + position encoding. where \( t \) represents the transformer encoder, \( N_e \) represents the number of entities, \( N_s \) represents the number of sentences.

(2) GAT Module: We construct a graph based on the built RST tree and use Graph Attention Network (GAT) \citep{veličković2018graph} to learn the information between different nodes, representing rich structural information between EDUs. \( E_g = [e_g^1, e_g^2, ...e_g^{N_e}]=\text{GAT}(\{n_1, n_2, ..., n_N\}) \). Similarly, \( C_g = [c_g^1,c_g^2, ..., c_g^{N_s}]=\text{GAT}(\{n_1, n_2, ..., n_N\})\). where g represents the GAT, \( n_i \) represents each node in the RST tree. Finally, \( e_g^i \) takes the node representation of the EDU where entity \( e_i \) is located, and \( c_g^i \) similarly takes the node representation of the EDU where sentence \( c_i \) is located.

The final entity representation is \( E = [e_t^1 \tensorconcat e_g^1, ..., e_t^{N_e} \tensorconcat e_g^{N_e}] \), \( C = [c_t^1 \tensorconcat c_g^1, ..., c_t^{N_s} \tensorconcat c_g^{N_s}] \).
where \( \tensorconcat \) represents concatenation operation. 

\subsubsection{Event Type Classification}
We perform max-pooling on the document-level encoding  \( C \) obtained from the previous step to get the document embedding \( d \). Then, we use a 3-layer Multi-Layer Perceptron (MLP) for event type classification. 
\begin{equation}
   P_t = \text{softmax}(\text{MLP}(d)) 
\end{equation}
\( P_t \) represents the probability of each event type. We select the event type corresponding to the highest probability as the final event type.

\subsubsection{Event Role Extraction}
We follow the approach of \citet{zheng-etal-2019-doc2edag} to construct an entity-based directed acyclic graph (EDAG) from the table event records. For each event type, we manually define the sequence of event roles. Then, we transform each event record into a parameter chain list according to this sequence, where each parameter node is either an entity or a special empty parameter NA. By sharing the same prefix path, we merge these lists into the EDAG. We perform path extension on each leaf node of the EDAG. For each entity to be extended, we create a new node for the entity based on the current role and connect the leaf node with the new node. We implement path extension as a classification task.

\subsection{Role Hierarchy}
\label{role hierarchy}
To address potential conflicts in event roles during the information integration process, we constructed an event hierarchy for the nine event types to resolve conflicts. When the same entity holds multiple roles, we select the highest-level role as the final result. The specific information about the role hierarchy is shown in Table \ref{table.role hierarchy}, with levels ranging from Level 1 to Level 5 in decreasing order of priority.
For more details, please refer to our code repository.
\begin{table*}[!ht]
\centering
\resizebox{\textwidth}{!}{
\renewcommand{\arraystretch}{1.5}
\begin{tabular}{l|c|c|c|c|c}
\hline
 & {Level 1}   & {Level 2}    & {Level 3}   & {Level 4} & {Level 5} \\
\hline
{ATTACK} 
& \makecell{Attacker\\Victim\\Direct Target} & \makecell{Eyewitness\\First Responder} & \makecell{Emergency Service\\ Investigator\\Intelligence Analyst} & \makecell{Reporter\\Analyst\\Policy Maker}& \makecell{Bystander\\Commentator\\Academic Researcher}\\
\hline
{SPORT} 
& \makecell{Winner, MVP\\Loser} & \makecell{Coach, Referee\\ Key Player} & \makecell{Participant\\Team Doctor\\Tactical Analyst} & \makecell{Sponsor\\ Spectator\\Media}& \makecell{Security Personnel\\ Event Organizer\\Volunteer}\\
\hline
{UNK} 
& \makecell{Main Participants} & \makecell{Directly Affected} & \makecell{Recorders\\Witnesses} & \makecell{Analysts\\Commentators}& \makecell{Bystanders}\\
\hline
{ELECTION} 
& \makecell{Winning Candidate\\Losing Candidate\\Election Official} & \makecell{Voter\\Campaign Team-\\ Member\\Political Analyst} & \makecell{Observer\\Media\\Pollster} & \makecell{Supporter\\ Opponent\\Independent Commentator}&
\makecell{Security Personnel\\ Election Equipment-\\ Supplier\\Legal Advisor}\\
\hline
{GENERAL} 
& \makecell{Organizer\\ Keynote Speaker\\Sponsor} & \makecell{Participant\\ Volunteer\\Service Provider} & \makecell{Media\\ Security Personnel} & \makecell{Audience Membe\\ Commentator\\Industry Analyst}& \makecell{Remote Participan\\ Social Media-\\ Influencer\\Academic Researcher}\\
\hline
{DISASTER} 
& \makecell{Victim\\ Rescue Team\\Emergency Management-\\ Official} & \makecell{Medical Service-\\ Provider\\Volunteer\\Donor}  & \makecell{Analyst\\ Journalist\\International Aid-\\ Organization}& \makecell{Policy Maker}& \makecell{Observer\\ Commentator}\\
\hline
{ACCIDENT} 
& \makecell{Victim\\ At-Fault Party} & \makecell{Eyewitness\\ First Responder} & \makecell{Investigator\\ Legal Advisor} & \makecell{Media\\ Analyst}& \makecell{Bystander\\ Commentator}\\
\hline
{AWARD} 
& \makecell{Awardee\\ Nominee\\Presenter} & \makecell{Organizer\\ Judge\\Sponsor} &  \makecell{Attendee\\ Media\\Industry Analyst}& \makecell{Audience\\ Commentator\\Social Media-\\ Influencer}& \makecell{Security Personnel\\ Technical Support-\\ Staff\\Volunteer} \\
\hline
{OTHERS} 
& \makecell{Main Participants} & \makecell{Directly Affected} & \makecell{Supporters\\ Opponents} & \makecell{Observers\\ Recorders}& \makecell{Analysts\\ Commentators}\\
\hline
\end{tabular}
}
\caption{\label{table.role hierarchy}
Event Role Hierarchy, where each row represents a specific event type, is organized into five levels, with Level 1 being the highest and Level 5 being the lowest.
}
\end{table*}

\end{document}